\documentclass[conference]{IEEEtran}
\IEEEoverridecommandlockouts
\usepackage{cite}
\usepackage{amsmath,amssymb,amsfonts}
\usepackage{algorithmic}
\usepackage{graphicx}
\usepackage{textcomp}
\usepackage{xcolor}
\usepackage{graphicx}
\usepackage{amsmath}
\usepackage{diagbox}
\def\BibTeX{{\rm B\kern-.05em{\sc i\kern-.025em b}\kern-.08em
    T\kern-.1667em\lower.7ex\hbox{E}\kern-.125emX}}
\begin{document}

\title{Mini Lesions Detection on Diabetic Retinopathy Images via Large Scale CNN Features\\
}

\author
{\IEEEauthorblockN{Qilei Chen, Xinzi Sun, Ning Zhang, Yu Cao, Benyuan Liu}
\IEEEauthorblockA{\textit{Department of Computer Science} \\
\textit{University of Massachusetts Lowell}\\
\textit{Lowell, USA}\\
\{qilei\_chen, Xinzi\_Sun, ning\_zhang\}@student.uml.edu, \{ycao, bliu\}@cs.uml.edu}
%\and
%\IEEEauthorblockN{2\textsuperscript{nd} Xinzi Sun}
%\IEEEauthorblockA{\textit{Computer Science} \\
%\textit{The University of Massachusetts Lowell}\\
%Lowell, United States \\
%Xinzi\_Sun@student.uml.edu}
%\and
%\IEEEauthorblockN{3\textsuperscript{rd} Ning Zhang}
%\IEEEauthorblockA{\textit{Computer Science} \\
%\textit{The University of Massachusetts Lowell}\\
%Lowell, United States \\
%ning\_zhang@student.uml.edu}
%\and
%\IEEEauthorblockN{4\textsuperscript{th} Yu Cao}
%\IEEEauthorblockA{\textit{Computer Science} \\
%\textit{The University of Massachusetts Lowell}\\
%Lowell, United States \\
%ycao@cs.uml.edu}
%\and
%\IEEEauthorblockN{5\textsuperscript{th} Benyuan Liu}
%\IEEEauthorblockA{\textit{Computer Science} \\
%\textit{The University of Massachusetts Lowell}\\
%Lowell, United States \\
%bliu@cs.uml.edu}
%\and
}

\maketitle

\begin{abstract}
Diabetic retinopathy (DR) is a diabetes complication that affects eyes. DR is a primary cause of blindness in working-age people and it is estimated that 3 to 4 million people with diabetes are blinded by DR every year worldwide. Early diagnosis have been considered an effective way to mitigate such problem. The ultimate goal of our research is to develop novel machine learning techniques to analyze the DR images generated by the fundus camera for automatically DR diagnosis. In this paper, we focus on identifying small lesions on DR fundus images. The results from our analysis, which include the lesion category and their exact locations in the image, can be used to facilitate the determination of DR severity (indicated by DR stages). Different from traditional object detection for natural images, lesion detection for fundus images have unique challenges. Specifically, the size of a lesion instance is usually very small, compared with the original resolution of the fundus images, making them diffcult to be detected. We analyze the lesion-vs-image scale carefully and propose a large-size feature pyramid network (LFPN) to preserve more image details for mini lesion instance detection. Our method includes an effective region proposal strategy to increase the sensitivity. The experimental results show that our proposed method is superior to the original feature pyramid network (FPN) method and Faster RCNN.
\end{abstract}


\section{Introduction}
Diabetic Retinopathy (DR) is a leading problem of ophthalmic disease globally 
%, especially in fast developing countries such as China \cite{zhang2017prevalence}
and one of the most common complications of diabetes. The prevalence of DR in diabetic populations is as high as 24.7-37.5\% \cite{danaei2011national}.
%Without early diagnosis and therapy, 
People suffering from diabetes have a higher risk of developing DR as the elevated glucose levels can damage the retina blood vessels.
Early screening and regular checkup with the fundus camera have been reported as an effective approach to reduce the risk of blindness \cite{zhang2017prevalence}. Therefore, it is important to develop an effective tool for DR detection in early screenings to improve the healthcare outcome.

%In this paper we investigate the lesion detection problem in DR fundus image for automatic DR diagnositcs.

\begin{figure}
\centering
\includegraphics[width=0.7\columnwidth,height=\textwidth,keepaspectratio]{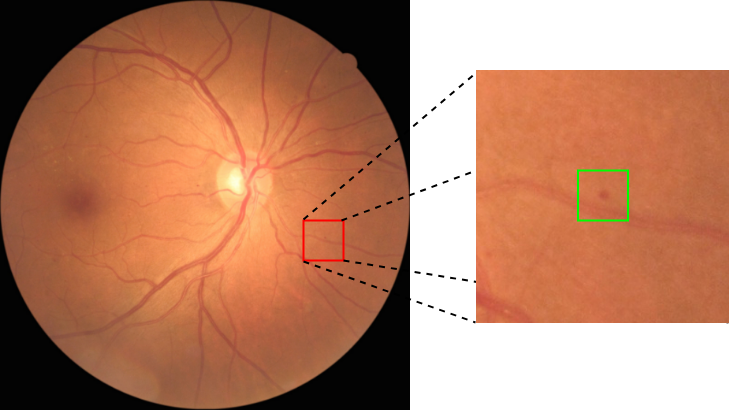}\caption{Mini lesion with zoom-in view.}\label{fig1}
\end{figure}

International Clinical Diabetic Retinopathy and Diabetic Macular Edema Disease Severity Scales \cite{wilkinson2003proposed} is a worldwide-used standard for severity diagnosis based on DR images. There are 5 stages for DR in the standard and the classification of the 5 stages is based on the location and number of the following 10 lesion categories: 1) blot hemorrhages, 2) micro-aneurysms, 3) hard exudate, 4) cotton wool spot, 5) fibrous proliferation, 6) venous beading, 7) intraretinal microvascular abnormity (IRMA), 8) neovascularization, 9) vitreous hemorrhage, 10) venous loop \cite{wang2018diabetic}. Normally, DR images need to be in a high-resolution format, so that the lesions can be shown in detail (see Fig.~\ref{fig1}), especially for the small size lesion categories 1)-4). It is a labor extensive job for an ophthalmologist to examine the whole image and figure out all the numbers, categories and locations of these lesions. In fact, finding lesions in DR images can be considered as an instance level multi-label visual object detection task and it has become a hot research topic in recent years \cite{dai2017retinal,wang2017zoom,yang2017lesion,zhao2018uniqueness}. 
%Many traditional methods have been proposed to handle the issue, such as HOG \cite{}, SIFT \cite{}, saliency \cite{} etc.
However, most of the previous methods are designed to detect a single lesion category (e.g., \cite{dai2017retinal,yang2017lesion}), or to detect multiple lesion categories in a sequence of steps \cite{zhao2018uniqueness}. Also, due to the lack of datasets with detailed location information for lesion instances, previous weakly supervised methods \cite{wang2017zoom} can only find suspicious lesion regions rather than fine grained individual lesion instances. In this study, we aim to develop a single model to detect all lesion instances of different categories in one round.

Traditional Faster-RCNN and FPN methods are not suitable for small lesion detection in our dataset, as the classification is based on the middle or top feature maps, which may loss the feature details of the small lesions. 
%Specifically, in FPN, the size of feature map for RoI-pooling and classification is one quarter of the original input size and it is even smaller for Faster-RCNN. Small size feature maps may lose the details of small objects, which can harm the identification of lesions with similar appearance in different categories.
To address this issue, we first propose a large-size feature pyramid network (LFPN) to be used in RCNN \cite{girshick2015fast} method. Our method increases the size of the bottom feature map to match that of the input image, which can better preserve the details of small targets on the original images, and is thus more effective for detecting small lesions. Furthermore, we find that normal region proposal strategy based on intersection-over-union (IoU) in region proposal network (RPN) tends to miss a lot of small true targets in our experiments. To alleviate this problem, we design an effective region proposal strategy,
%that includes small anchors during the training process to make the model more sensitive to small lesion targets.
in which small anchors containing center region of ground-truth will be set positive, to help RPN pay more attention to true mini lesion targets on the fundus image. 
The experiment results show that our methods can considerably improve the performance of lesion detection over the original Faster-RCNN and FPN methods. To the best of our knowledge, our work is the first effort for lesion detection at the instance level,  based on the unique DR image dataset we collected with lesion instances labeled by ophthalmologists. 

The major contributions of this work can be summerized as follows:
\begin{itemize}
    \item Firstly, we propose a large size feature map method in FPN to preserve details of high-resolution fundus images for detecting mini size lesions.
    \item Secondly, we design a center focus strategy in RPN to get more acceptable anchors for the lesion targets detection.
    \item Thirdly, our lesion detection method outperforms other state-of-art approaches in the experiments and provides a strong foundation for DR severity diagnosis. 
\end{itemize}
\section{Related Work}
In this section, we will review recent CNN-based object detection methods that will be used in our research. We also briefly introduce the International Clinical Diabetic Retinopathy and Diabetic Macular Edema Disease Severity Scales (ICDRDMEDSS) and show the important relationship between lesions detection and severity diagnosis. 
\subsection{CNN-based Object Detection}
With the explosive development of convolution neural networks (CNNs) these years, various CNN-based visual object detection methods such as Faster-RCNN \cite{ren2015faster}, SSD \cite{liu2016ssd} and YOLO \cite{redmon2016you} have been proposed. 
According to the process of detection, these methods can be devided in two categories: one-stage methods such as SSD, YOLO and two-stage methods such as Faster-RCNN. In one stage methods, object class classification and location regression are directly predicted through the feature map of backbone. Unlike one stage methods, the region proposal network (RPN), which can be described as a binary-label detector, is used in two stage methods as the first step. The binary-label detector can predict objectness and the positive results will be used in the second step for the object label classication and location regression. From previous studies, two-stage methods are often more effective for various size object detection. We use Faster-RCNN as the baseline in our experiments.
\subsection{Feature Pyramid Networks}
Image pyramids construction \cite{adelson1984pyramid} have been proven to be an effective method to handle the fundamental challenge of recognizing objects at vastly different scales in computer vision. With the prevelent of CNN methods, Lin et al.\cite{lin2017feature} proposed a Feature Pyramid Network (FPN) method built upon CNN features as a basic component in recognition systems for detecting objects at different scales. Faster-RCNN with FPN achieves better result on large scale natural object detection dataset COCO \cite{lin2014microsoft}. The standard FPN uses the last residual block of the 4 stages from the ResNets \cite{he2016deep} backbone as input and then goes through a top-down pathway to construct 4 feature layers, with the size ratio between adjacent layers set to be 2. In a standard FPN, the ratio of original image to the largest feature scale is 4. Larger size feature can preserve more details of the objects, which is especially important for small instances. Motivated by such intuition, we modify the standard FPN and extend the feature scale to the original size.  
\subsection{Diabetic Retinopathy Diagnosis}
\begin{figure}
\centering
\includegraphics[width=0.45\columnwidth,height=\textwidth,keepaspectratio]{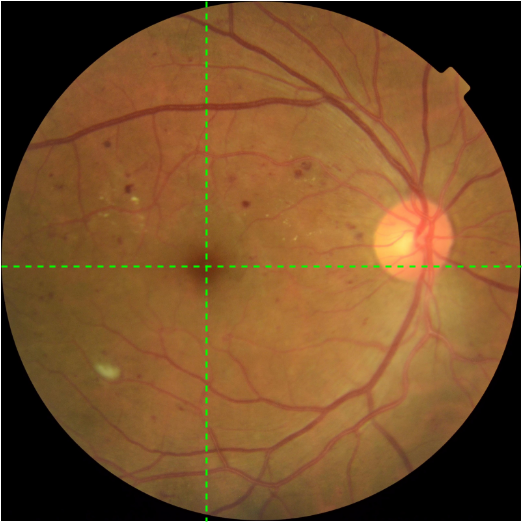}
\caption{DR image with 4 quadrants based on the center of marcula according to ICDRDMEDSS.}\label{fig5}
\end{figure}

International Clinical Diabetic Retinopathy and Diabetic Macular Edema Disease Severity Scales (ICDRDMEDSS) is proposed by Wilkinson in 2003 as one of the international standards, in which diabetic retinopathy can be classified into 5 stages: 1) No DR, 2) Mild non-proliferative DR, 3) Moderate non-proliferative DR, 4) Severe non-proliferative DR, 5) Proliferative DR. The stage diagnosis is based on the category and location of lesions on the fundus image. For example, having more than 20 hemorrhages in each of 4 quadrants without signs of proliferative retinopathy is the indicative condition of the 4th stage. Quadrants are centered by macula (shown in Fig~\ref{fig5}). EyePACS \cite{cuadros2009eyepacs} is a well-known large scale dataset for DR Diagnosis in ICDRDMEDSS standard. It contains more than 100,000 fundus images and each image was labeled with an integer ranging from 0 to 4, indicating the stage of DR. In EyePACS, DR diagnosis is considered as a task of image classification. In fact, location and classification of lesions on fundus can show detail of DR diagnosis. Therefore lesion instance detection can provide a more effective way to assist ophthalmologists to view the condition of DR.

\section{Dataset and Method}
%Just like most of CNN-based methods that are deiven by large scale dataset
Data collection is a vital step for lesion detection methods based on convolutional neural network (CNN). We developed our own labeling tool, which can be easily applied to mark the bounding box and category of lesion instances in an image. %Another important step is to analysis the special features of the dataset, the analysis result will be shown in the following section.
After the manual labeling step, we tally the number of object instances and compute the area ratio for each lesion category. The summary information will be shown in Section $III.A$. LFPN is designed for preserving and making full use of the small size object features. The most important part of LFPN is that we use the stride-1 feature map, which is the same size as the input image, for RoI-pooling and classification in RCNN (shown in Fig. \ref{fig2}). This layer can preserve the details of small object features as much as possible. Another difference from FPN is that we use the top layer for region proposal and then map the region-of-interests (ROIs) to the stride-1 feature map. Details of LFPN can be seen in Fig. \ref{fig2}.
%, which is sufficient for targeting all the lesion instances and saves computing resource.
For small target proposal, IoU is not an effective metric to locate the lesion targets, especially when the object constitates only a small part in the center of the ground-truth bounding box. We add a center-focused condition in the proposal strategy for the situation mentioned above, which is shown to improve the result in the experiments.
\subsection{Dataset Analysis}

The dataset contains $5,198$ images with a resolution of $2136\times3216$, including fundus pictures from 500 patients and covering all 5 severity stages. All the original images were preprocessed to remove the left and right hand side black parts with low pixel values. This helps the model focus on the fundus part with a new size of $2136\times2136$. There are 10 lesion labels as mentioned in the introduction part in our labeling tool and a flexible bounding box tool is provided to mark the location and category of each lesion. Each annotation box represents a single lesion instance and contains 5 values $(x,y,w,h,c)$, where $(x,y)$ is the coordinate in a fundus image for the upper-left corner of ground-truth, $(w,h)$ is the size of the box and $c$ is the label of the lesion. The dataset was randomly divided into 4 equal parts and each part was handled by one specialist and then validated by 3 other specialists.

\begin{table}
\caption{Summary of lesions in 10 different categories. Labels correspond to the categories description order mentioned in the introduction.}\label{tab1}
\centering
\begin{tabular}{|c|p{1cm}<{\centering}|p{1cm}<{\centering}|p{1cm}<{\centering}|p{1cm}<{\centering}|}
\hline
\diagbox {Label}{Set} & Total & Train & Validation \\
\hline
1& 18493 & 14720 & 3773 \\
\hline
2& 7703 & 6301 & 1402 \\
\hline
3& 9316 & 7403 & 1913 \\
\hline
4& 654 & 537 & 117 \\
\hline
5& 34 & - & - \\
\hline
6& 15 & - & - \\
\hline
7& 25 & - & - \\
\hline
8& 49 & - & - \\
\hline
9& 14 & - & - \\
\hline
10& 1 & - & - \\
\hline
\end{tabular}
\end{table}

\begin{table}
\center
\caption{The average lesion-to-image ratios for categories 1)-4).}\label{tab2}
\begin{tabular}{|c|c|c|c|c|}
\hline
Label&1&2&3&4 \\
\hline
Ratio&0.07244\%&0.05390\%&0.31672\%&0.23976\%\\
\hline
\end{tabular}
\end{table}

Table~\ref{tab1} summarizes the number of instances for each lesion category. We can see that categories 1)-4) make up the majority $(99.3\%)$ of the dataset. 
%That makes sense because these four categories are indicators of the first three stages of severity \cite{}, which are more valuable for diagnosis and treatment.
These four categories are more valuable for early diagnosis because they are indicators for the frist three stages of severity \cite{wang2018diabetic}. Compared with the other 6 categories of lesion, labeling of categories 1)-4) are more challenging and labor-consuming as the area ratios are very small on a fundus image. The average lesion-to-image ratios for categories 1) - 4) are listed in Table~\ref{tab2}. % The number of area ratio distribution can be seen in Table~\ref{tab2} or supplementary part.

\subsection{Large Size Feature Pyramid Network}\label{AA}
\begin{figure}
\centering
\includegraphics[width=0.75\columnwidth,height=\textwidth,keepaspectratio]{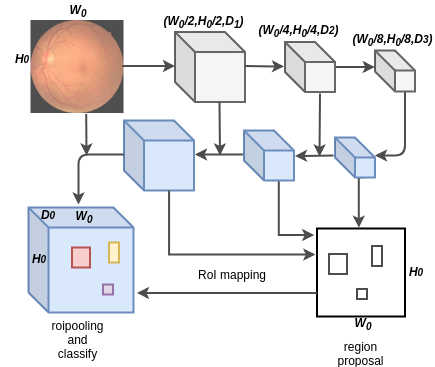}
\caption{The architecture of large-size feature pyramid network (LFPN). $(W_0,H_0)$ is the input image size and $D_i$ is the channel size of feature maps.}\label{fig2}
\end{figure}

Fig.~\ref{fig2} illustrates the architecture of our proposed large-size feature pyramid network (LFPN). As in FPN, we upsample the spatial resolution of the feature map by a factor of 2 and then merge each upsampled map with the corresponding bottom-up map. The difference is that we will continue to upsample the feature map until it reaches the input size ($W_{0}$, $H_{0}$). The input image will be considered as a feature map and it will go through a $1\times1\times D_0$ convolutional layer to increase the channel dimension from $3$ to $D_0$. After that, the new $D_0$-dimension layer will be merged with the upsampled feature map with size ($W_{0}$, $H_{0}$) by operation of elementwise-sum. For example, if Resnet-101 is used as the backbone, we will get a set of feature maps containing 6 layers \{$P_{0}$, $P_{1}$, $P_{2}$, $P_{3}$, $P_{4}$, $P_{5}$\}, where layer $P_{0}$ has the same size as intput data, while FPN only contains $P_2$ to $P_5$.
With a large feature map size, $P_{0}$ can sufficiently preserve the feature of small lesions. In FPN, every feature layer is used for both region proposal and class prediction. In LFPN, we only use the smaller size feature maps, i.e. $P_{1-5}$, as the region proposal layers, and then the proposed regions of interest (RoI) will be mapped to layer $P_{0}$. Unlike the application condition of FPN, in which layers $P_{1-4}$ have to be presented for multi-scale \cite{xiao2016multi} objects, our approach only concentrates on small-scale lesion targets and layers $P_{1-4}$ act as linking chain and feature producers. For a GPU with fixed memory capacity, LFPN can provide larger $P_{0}$ size and better result than FPN. Compared with faster-RCNN, the region proposal layer remains the same, but our RoI pooling is performed on the largest layer rather than the top feature layer of the backbone, thus the classifier can use more detailed features to get better results.
\subsection{Center-Focus Target Proposal}
Region proposal network (RPN) \cite{ren2015faster} is a vital part in RCNN methods as it predicts object bounds and objectness scores at each position on the feature map. In RPN, an anchor $a$ is centered at the sliding window in question and is associated with a scale and aspect ratio \cite{ren2015faster}. The region proposal part will assign an anchor with a positive label if it matches ether one of conditions mentioned in \cite{ren2015faster} and with a negative label if the IoU ratio between the anchor and all ground-truth boxes is lower than 0.5. 

\begin{figure}
\centering
\includegraphics[width=0.7\columnwidth,height=\textwidth,keepaspectratio]{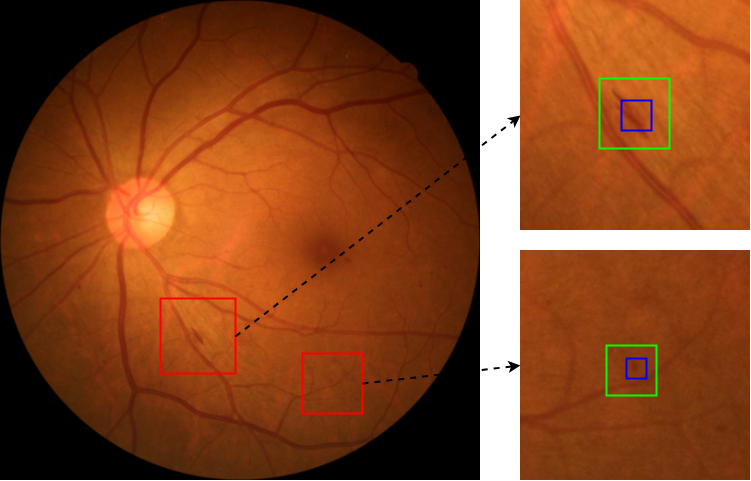}
\caption{Red rectangle parts of the left image are zoomed in and shown on the right. Green rectangles are the ground-truth boxes and blue ones are proposal targets.}\label{fig3}
\end{figure}

In fact, manual labeling of small size lasions is a difficult task even for specialists and usually the bounding box contains part of lesion region rather than strictly flollowing along its contour (see green boxes in Fig.~\ref{fig3}). The center region are the most important part of the bounding box when the area ratio of the lesion to the ground-truth box is less than 0.5. The region proposal strategy in \cite{ren2015faster} may reject some acceptable anchor boxes. For example, in Fig.~\ref{fig3}, the small blue anchor fits the whole lesion region perfectly but the IoU ratio is less than 0.5 and the anchor will be set to negative based on the proposal condition mentioned in \cite{ren2015faster}.
In addition to the original condition, we introduce a center-focus (CF) target proposal condition to help mitigate the problem of rejecting suitable anchors:

\begin{equation}
f_a(IoU,C_a^g)=
\begin{cases}
1,&\text{(IoU/$A_a^g$$>$0.1$\land$$C_a^g$=1)}\\
    &\text{$\lor$(IoU/$A_a^g$$\geq$0.5)}\\
0,&\text{IoU/$A_a^g$$<$0.5}
\end{cases}
\label{e1}
\end{equation}
\begin{equation}
C_a^g=
\begin{cases}
1,&\text{if $c_g$ $\in$ a}\\
0,&\text{otherwise,}
\end{cases}
\label{e2}
\end{equation}
      \\
where $A_a^g$ represents the merged area of ground truth $g$ and anchor $a$, $C_a^g$ is 1 when the center of ground truth resides in the anchor and $A_a$ represents the area of anchor $a$. Equation $(\ref{e2})$ presents that if the anchor contains the center region of the ground-truth box and equation $(\ref{e1})$ means that the anchor with IoU ratio greater than 0.5 or location near the center of ground-truth is positive. %Both equation $(\ref{e1})$ and $(\ref{e2})$ together determine whether an anchor is positive. 

\section{Experiments}

In this section, we will first introduce a new criterion in our experiments and then describe the details of hyperparameter settings and analysis of the results. We select all images with small lesion categories $1)-4)$ as the experimental dataset. During the experiments, we randomly divide the dataset into 2 parts: one is for training while the other for validation with a 4:1 ratio. The number of lesion instances of both sets are shown in the last two rows of Table~\ref{tab1}. %Manual labeling is a process of sampling label, which means that not all the lesions are marked in the dataset. In our experiments, we focus on the sensitivity according to the IoU between ground truth and prediction result during the validation step, rather than the accuracy because a lot of unlabeled lesion well be detected, which is more resonable for the dataset.
Manual lesion labeling is a challenging task, and some lesions may be not marked. Our CNN-based detector can even detect the lesion that actually exist but are not marked in ground-truth. It is more reasonable to use the sensitivity of results as the quality metric on the validation set.
In our experiments, horizontal flipping is applied during the training for data augmentation. LFPN is implemented on MXnet and the experiments are performed on 4 NVIDIA TESLA K80 GPUs.

\subsection{Center-Focus Criterion}
Standard criterion for natural object detection is based on IoU ratio only \cite{ren2015faster}, which means the IoU ratio between a true positive object and the ground-truth should be above the threshold 0.5. In our DR dataset, the bounding box of ground-truth lesion instance contains a lot of context as the green rectangle shown in Fig.~\ref{fig3} and the center region of ground-truth are the main information of lesions. A predicted object thar has IoU ratio less than 0.5 but contains center region is still acceptable in our application. Corresponding to the CF target proposal condition mentioned in Section $III$, we develop our own center-focus criterion. 
In this criterion, a prediction rectangle should be positive when the IoU ratio is more than 0.1 and the rectangle contains the center point of ground-truth.

\subsection{LFPN for Lesion Detection}
We train LFPN-RCNN on our dataset and use another two methods for baselines, namely Faster-RCNN and Faster-RCNN with FPN. The same parameter settings are used for all three methods. We apply Resnet101 as the backbone, which has been pretrained on Imagenet \cite{deng2009imagenet}. The input size of the network for the three methods mentioned above is $1120\times1120$. For LFPN, layers $P_{1-5}$ are applied for region proposal. %The size of layer $P_5$ is $17\times17$ thus the number of boxes proposed by RPN can reach $17\times17\times21$, which is enough for all the lesion instances in one image.
The base anchor number of each location in one layer is $21$. In our work, we use a small anchor scale list $\{0.02,0.05,0.1,0.2,0.5,1,2\}$ for all the three methods to fit the small lesion detection.
\begin{table}
\caption{Summary of sensitivities on 4 categories. The last three rows show the results with center-focus (CF) proposal strategy.}\label{tab3}
\centering
\begin{tabular}{|c|p{1cm}<{\centering}|p{1cm}<{\centering}|p{1cm}<{\centering}|p{1cm}<{\centering}|}
\hline
\diagbox{Method}{Category} & Blot Hemorrhages & Micro-aneurysms & Hard Exudate & Cotton-wool Spot \\
\hline
FasterRCNN& 84.63\% & 80.98\% & 85.85\% & 72.22\% \\
\hline
FPN& 90.21\% & 88.10\% & 91.79\% & 74.60\% \\
\hline
\textbf{LFPN}&  \textbf{92.76\%} &  \textbf{91.96\%} & 90.51\% &  \textbf{75.40\%} \\
\hline
FasterRCNN+\textbf{CF}& \textbf{87.58}\% & \textbf{81.37}\% & 89.23\% & 70.63\% \\
\hline
FPN+\textbf{CF}& \textbf{91.90\%} & \textbf{89.02\%} & 88.46\% & 62.70\% \\
\hline
\textbf{LFPN+CF}&  \textbf{93.01\%} &  86.26\% & \textbf{93.79}\% &  \textbf{79.73\%} \\
\hline
\end{tabular}
\end{table}
During the validation step, we set the max prediction number of one image as 100 and the confidence score threshold is 0.1. The results of sensitivity with CF criterion is shown in the first 3 rows of Table.~\ref{tab3}. For the lesion categories of 1), 2) and 4), the results of our proposed LFPN are superior to the other two methods, which means that the LFPN is more suitable for smaller lesions detection as lesions of categories 1) and 2) have a lower area ratio on average.

\subsection{Detection with Center-Focused Target Proposal}
\begin{figure}
\centering
\includegraphics[width=0.8\columnwidth,height=\textwidth,keepaspectratio]{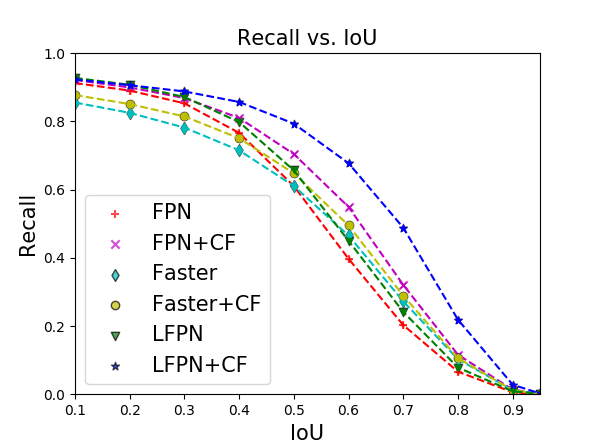}
\caption{Recall $vs.$ IoU overlap ratio.}\label{fig4}
\end{figure}

Fig.~\ref{fig4} show the recall results for ablations of the design choices for CF proposal. The blue line represents the result from LFPN with CF proposal strategy, which can help the network focus more on the center of the main feature of each ground truth. We observe that the recall is still very high even when the IoU ratio threshold increases to 0.6. From the last three rows of Table.~\ref{tab3}, we can see that our CF strategy is effective on most results of the lesion categories. The result of micro-aneurysms worsens with the CF strategy. We analyze the result for each lesion category and find that 1) blot hemorrhages and 2) micro-aneurysms have high cross-mistakes. Actually, some small hemorrhages are similar to micro-aneurysms on appearance. With the increasing accuracy of hemorrhages, the classifier becomes more sensitive to hemorrhages, which may harm the recognition ability of micro-aneurysms. We will follow up with the issue in our future research.
\section{Conclusion}
In this work, we propose LFPN to detect small lesion instances on DR images. The proposed architecture has two advantages: Frist, we use large CNN feature maps, which has the same size as the input image and contains detail of small lesion features, and thus is more effective for object classification in the second stage of RCNN. Another is that we use the top layer for region proposal, which is computing resource efficient. To enhance the target presentation, the center-focused condition is applied to the proposal strategy. In fact, small lesions detection in large size images is a common problem for medical image processing. Our method can automatically propose a lot of useful lesion regions on large medical images, thus improves doctor diagnosis accuracy and efficiency and saves medical resources.

\bibliographystyle{./bibliography/IEEEtran}
\bibliography{ref}

\end{document}